%% file: main.tex
\documentclass[conference]{IEEEtran}
\IEEEoverridecommandlockouts
\usepackage{latexsym}
\usepackage{graphicx}
\usepackage{amsfonts,amssymb,amsmath}
\def\BibTeX{{\rm B\kern-.05em{\sc i\kern-.025em b}\kern-.08em T\kern-.1667em\lower.7ex\hbox{E}\kern-.125emX}}

\usepackage{algorithmic}
\usepackage[ruled,linesnumbered]{algorithm2e}
\usepackage{siunitx}
\usepackage{todonotes}
\usepackage{graphicx}
\usepackage{subfigure}
\usepackage[font=footnotesize,skip=10pt]{caption}
\usepackage{url}

\usepackage[acronym]{glossaries}
\newcommand{\newac}{\newacronym}

\makeglossaries

\newac{los}{\text{LoS}}{line-of-sight}
\newac{nlos}{\text{NLoS}}{non-line-of-sight}

\input{math_lib}
\usepackage{anyfontsize}

\begin{document}

\title{
Model-aided Federated Reinforcement Learning for Multi-UAV Trajectory Planning in IoT Networks
}

\author{
\IEEEauthorblockN{Jichao Chen$^{1,2}$, Omid Esrafilian$^{1}$, Harald Bayerlein$^{2}$, David Gesbert$^{1}$, and Marco Caccamo$^{2}$}
\IEEEauthorblockA{$^{1}$Communication Systems Department, EURECOM, Sophia Antipolis, France\\
$^{2}$TUM School of Engineering and Design, Technical University of Munich, Germany\\
\text{\{jichao.chen, h.bayerlein, mcaccamo\}@tum.de}, \{omid.esrafilian, david.gesbert\}@eurecom.fr}
}

\maketitle

\begin{abstract}  

Deploying teams of unmanned aerial vehicles (UAVs) to harvest data from distributed Internet of Things (IoT) devices requires efficient trajectory planning and coordination algorithms. Multi-agent reinforcement learning (MARL) has emerged as a solution, but requires extensive and costly real-world training data. To tackle this challenge, we propose a novel model-aided federated MARL algorithm to coordinate multiple UAVs on a data harvesting mission with only limited knowledge about the environment. The proposed algorithm alternates between building an environment simulation model from real-world measurements, specifically learning the radio channel characteristics and estimating unknown IoT device positions, and federated QMIX training in the simulated environment. Each UAV agent trains a local QMIX model in its simulated environment and continuously consolidates it through federated learning with other agents, accelerating the learning process. A performance comparison with standard MARL algorithms demonstrates that our proposed model-aided FedQMIX algorithm reduces the need for real-world training experiences by around three magnitudes while attaining similar data collection performance.

\end{abstract}

\input{sections/01intro}

\input{sections/02system}
\input{sections/03mdp}

\input{sections/04model}
\input{sections/05results}

\section{Conclusion}\label{sec:conclusion}
We have proposed a novel model-aided FedQMIX algorithm for designing cooperative multi-UAV trajectories in data harvesting missions. By leveraging federated learning and training the QMIX model within a learned simulation environment, our approach has significantly accelerated the learning process and reduced the extensive requirement for real-world experiences while achieving the same data collection performance as the baseline methods. In future work, we plan to fully decentralize the environment model learning approach.

\section*{Acknowledgments}
This work was partially funded by the HUAWEI France supported Chair on Future Wireless Networks at EURECOM, and by the German-French Academy for the Industry of the Future, under project 3CSI. Marco Caccamo was supported by an Alexander von Humboldt Professorship endowed by the German Federal Ministry of Education and Research.

\bibliographystyle{IEEEtran}
\bibliography{literature.bib}

\end{document}

%% file: math_lib.tex
\DeclareMathAlphabet{\mathbit}{OML}{cmr}{bx}{it}
\DeclareMathAlphabet{\mathsf}{OT1}{cmss}{m}{n}
\DeclareMathAlphabet{\mathTXf}{OT1}{cmss}{bx}{it}

\DeclareMathOperator{\SNR}{SNR}
\DeclareMathAlphabet{\mathpzc}{OT1}{pzc}{m}{it}



%% file: sections/01intro.tex
\section{Introduction}\label{sec:Intro}

Unmanned aerial vehicles (UAVs) have attracted growing interest in communication networks due to their high mobility, flexibility, and ease of deployment \cite{luo2023path}. One of the most important applications is data harvesting from geographically dispersed Internet of Things (IoT) sensor devices, where UAVs acting as mobile base stations, can establish efficient line-of-sight (LoS) links with IoT devices. The performance of data harvesting depends on the dynamic trajectories of UAVs, and the behavior of each UAV can affect the tasks of other UAVs. Therefore, it is crucial to design effective trajectories for UAVs, while considering cooperation between them, in order to ensure efficient data collection.

Recently, deep reinforcement learning (DRL) algorithms have been extensively employed to design UAV trajectories for communication. The authors in \cite{samir2020age} employ a centralized deep deterministic policy gradient (DDPG) approach to optimize multi-UAV trajectories, aiming to minimize the age of information (AoI) during a data collection mission. However, centralized learning leads to scalability issues with increasing agent numbers. The authors in \cite{bayerlein2021multi} avoid this issue by utilizing a double deep Q-network (DDQN) approach based on centralized learning with decentralized execution (CLDE) for multi-UAV path planning to maximize collected data from IoT sensor nodes. The approach is focused on adapting to a wide range of scenario parameters, but incurs a high training cost. QMIX, a classic multi-agent reinforcement learning (MARL) algorithm with a nonlinear value decomposition that we also base our approach on, is employed in \cite{wang2023cooperative} for multi-UAV trajectory planning to minimize the average AoI, outperforming other baselines like cluster-based and independent DQN-based approaches. Moreover, the authors in \cite{wang2023ensuring} develop a fully decentralized MARL framework to maximize data collection while minimizing AoI and maintaining a certain AoI threshold. This method utilizes a transformer for temporal modeling and introduces an intrinsic reward mechanism to enhance the exploration, achieving better results than other classic MARL algorithms. In contrast to our approach, all mentioned works have large training data requirements and assume full prior knowledge of all device or user positions.

In our work, we also make use of the idea of federated reinforcement learning (FRL) where learned information is exchanged between agents without uploading raw collected data to a central server, thereby accelerating learning without incurring too high communication overhead costs \cite{qi2021federated}. For instance, the authors in \cite{shahbazi2022federated} utilize FRL to design multiple UAV trajectories for user localization, lowering localization error and increasing convergence speed. In another related application, the authors in \cite{wu2022distributed} propose a distributed federated multi-agent DDPG algorithm to optimize trajectories for air and ground unmanned vehicles in emergency situations leveraging FRL to address data isolation and to accelerate the convergence. 

Despite the satisfactory performance of the aforementioned algorithms in their respective domains, DRL-based trajectory planning algorithms may lose their benefits in real-world scenarios, since collecting real-world training data requires expensive interaction with actual physical systems \cite{dulac2019challenges}. Our proposed solution to this challenge is based on \cite{esrafilian2021model}, a first attempt to design a model-aided DRL algorithm for UAV data harvesting, however only for the single UAV case in a simpler environment containing devices with unlimited data.

In this paper, we consider a scenario where multiple energy-limited UAVs cooperate to collect data from ground devices with only limited data volume. In our approach\footnote{The final version of this paper has been accepted by IEEE GLOBECOM Workshops 2023. Code available at \url{https://github.com/Cirrick/Multi_UAV_Data_Harvesting}.}, we make use of the popular QMIX algorithm \cite{rashid2020monotonic} and the idea of FRL \cite{qi2021federated}. To the best of our knowledge, this is the first work that employs FRL with a learned environment model to design multi-UAV trajectories for IoT data collection. Our main contributions are summarized as follows:
\begin{itemize}
    
    \item We exploit measurements collected by the UAVs to learn a digital twin of the real-world environment suitable for training MARL algorithms. To this end, we estimate unknown IoT device locations using particle swarm optimization (PSO) without requiring prior knowledge of the radio channel characteristics.
    
    \item Leveraging the learned environment, we introduce a model-aided QMIX-based algorithm to solve the multi-UAV data harvesting problem that reduces the amount of costly real-world training data by around three magnitudes compared to standard MARL algorithms.  
    
     \item We further propose a model-aided federated QMIX (FedQMIX) algorithm to enhance convergence speed through federated learning consolidating the locally trained QMIX models at each UAV. This approach helps to distribute the computational load over all UAVs and to utilize all available resources efficiently. 
     
\end{itemize}

%% file: sections/02system.tex
\section{System Model and Problem Formulation}\label{sec:SysModel}
We consider an urban environment where a set of UAVs $\mathcal{I}=\{1,\cdots,I\}$ are deployed to collect data from various stationary ground IoT devices. The mission time duration is discretized into $T$ equal time slots with length $\Delta t$, which are chosen sufficiently short so that the velocity of each UAV can be assumed to remain constant within a single time slot. The position of the $i$-th UAV at time step $t$ is denoted by $\mathbf{p}_{t}^{i} = [x_{t}^{i}, y_{t}^{i}, h^{i}]^{\operatorname{T}} \in \mathbb{R}^{3}$, $t\in [0, T]$, where $h^{i}$ represents the altitude of the $i$-th UAV. We assume that each UAV flies at a different altitude for collision avoidance and that each UAV's altitude remains constant throughout mission duration. Each UAV flies from a predefined starting point $\mathbf{p}_{I}$ and will return to a terminal point $\mathbf{p}_{F}$ at the end of the mission. UAVs do not collect data from the devices at the final time step $T$.

A set of ground devices $\mathcal{U}=\{1,\cdots,K\}$ is distributed in the environment, with the $k$-th device located at $\mathbf{u}^{k} = [x^{k}, y^{k}, 0]^{\text{T}}\in\mathbb{R}^{3}$. For device $k$, the amount of remaining data in its buffer at time step $t$ is denoted by $D^k_t$. The buffer of each device is initialized as $D^k_0=D^k_{init}$ at the beginning of the mission. Moreover, the ground devices are divided into two distinct groups: devices with known locations (referred to as \emph{anchor devices}) $\mathcal{U}_{known}$, and devices with unknown locations $\mathcal{U}_{unknown}$. Accordingly, we have $\mathcal{U}=\mathcal{U}_{known} \cup \mathcal{U}_{unknown}$.

\subsection{UAV Model}
The action space of each UAV is defined as 
\begin{equation}
\setlength\abovedisplayskip{5pt}
\setlength\belowdisplayskip{5pt}
\mathcal{A}=\left\{\underbrace{\begin{bmatrix}0\\0\\0\end{bmatrix}}_{\text{hover}},\underbrace{\begin{bmatrix}0\\c\\0\end{bmatrix}}_{\text{north}},\underbrace{\begin{bmatrix}-c\\0\\0\end{bmatrix}}_{\text{west}},\underbrace{\begin{bmatrix}0\\-c\\0\end{bmatrix}}_{\text{south}},\underbrace{\begin{bmatrix}c\\0\\0\end{bmatrix}}_{\text{east}},\underbrace{\begin{bmatrix}0\\0\\0\end{bmatrix}}_{\text{no-op}}\right\}, \label{eq:Action space}
\end{equation}
where $c$ denotes the distance that the UAV can move within a single time step. The no-op action refers to no operation, i.e., no movement and no energy consumption, which can only be chosen when the UAV has drained its energy. Given the executed action $\mathbf{a}_{t}^{i} \in \mathcal{A}$, the position of the UAV evolves as
\begin{equation}
\setlength\abovedisplayskip{3pt}
\setlength\belowdisplayskip{3pt}
    \mathbf{p}_{t+1}^{i}=\mathbf{p}_{t}^{i}+\mathbf{a}_{t}^{i}.
    \label{eq:UAV_mobility}
\end{equation}
\noindent
We assume the hover action consumes half the energy of the movement action as in \cite{esrafilian2021model}. Denoting the remaining battery of the $i$-th UAV at time step $t$ by $b_{t}^{i}\in\mathbb{R}$, which evolves according to 
\begin{equation}
\setlength\abovedisplayskip{2pt}
\setlength\belowdisplayskip{2pt}
    b_{t+1}^{i}=\begin{cases}
        b_{t}^{i},& \text{if } \mathbf{a}_{t}^{i}={\text{no-op}},\\
        b_{t}^{i}-0.5,& \text{if }\mathbf{a}_{t}^{i}={\text{hover}},\\
        b_{t}^{i}-1,&\text{otherwise}.
    \end{cases}
    \label{eq:UAV_battery}
\end{equation}
The data harvesting mission terminates when all UAV batteries are depleted.

\subsection{Channel Model}
The channel gain between UAV $i$ and device $k$ at time step $t$ is modeled as \cite{esrafilian2021model}
\noindent
\begin{equation}
\setlength\abovedisplayskip{3pt}
\setlength\belowdisplayskip{3pt}
    g_{t}^{i,k}\!=\!\left\{\begin{array}{ll}
    \beta_{\text{LoS}}+\alpha_{\text{LoS}} \log_{10}(d_{t}^{i,k})+\eta_{\text{LoS}}, & \small{\text{if} \text{ LoS}}, \\
    \beta_{\text{NLoS}}+\alpha_{\text{NLoS}} \log_{10}(d_{t}^{i,k})+\eta_{\text{NLoS}}, & \small{\text{if} \text{ NLoS}},
    \end{array}\right.
\end{equation}
\noindent
where $d_{t}^{i, k}=\left\|\mathbf{p}_{t}^{i}-\mathbf{u}^{k}\right\|_2$ is their absolute distance. Let $z\in\{\text{LoS, NLoS}\}$ denote either a LoS or non-LoS (NLoS) condition, $\alpha_{z}$ is a path loss constant, $\beta_{z}$ is the average channel gain at the reference distance $d_{0}=1\text{m}$, and $\eta_{z}$ represents the shadowing component modeled as a Gaussian distribution $\mathcal{N}(0,\sigma_{z}^2)$. Note that both the channel model and associated parameters are unknown and need to be learned by the UAVs. The signal-to-noise ratio (SNR) is given by
\begin{equation}
\setlength\abovedisplayskip{2pt}
\setlength\belowdisplayskip{2pt}
    \SNR_{t}^{i,k} = \frac{P10^{0.1g_{t}^{i,k}}}{\sigma^2}, \label{eq:SNR}
\end{equation}
where $P$ is the transmit power, and $\sigma^2$ is the white Gaussian noise power at the receiver. Additionally, the channel gain and SNR between two UAVs can be modeled analogously. Furthermore, the information rate is given by $R_{t}^{i,k} = \log_{2}(1 + \SNR_{t}^{i,k})$ assumed to be constant within each time slot. 

\subsection{Channel Access Protocol} \label{sec:Channel Access Protocol}
We assume that the communication between the UAVs and ground devices follows a time-division multiple access (TDMA) approach. Denoting the data collection status by $q_{t}^{i,k}\in \{0,1\}$, $q_{t}^{i,k} = 1$ indicates the $i$-th UAV collects data from the $k$-th device at time step $t$, and otherwise $q_{t}^{i,k} = 0$. We assume that each UAV can collect data from only one device at each time step, imposing the following constraint
\begin{equation}
\setlength\abovedisplayskip{2pt}
\setlength\belowdisplayskip{2pt}
    \sum_{k=1}^{K}\ q_{t}^{i,k} \leq 1,\quad \forall i\in \mathcal{I},\ t\in [0, T-1].
    \label{eq:device}
\end{equation}
Similarly, every device's data can only be collected by a single UAV, i.e.,
\begin{equation}
\setlength\abovedisplayskip{2pt}
\setlength\belowdisplayskip{2pt}
    \sum_{i=1}^{I}\ q_{t}^{i,k} \leq 1,\quad \forall k\in \mathcal{U},\ t\in [0, T-1].
    \label{eq:UAV}
\end{equation}
The value of $q_{t}^{i,k}$ is set according to the max-rate rule in \cite{bayerlein2021multi}: among all the ground devices, only the device with available data and the highest $\SNR_{t}^{i,k}$ is considered for data collection by the $i$-th UAV. 
If the $i$-th UAV collects data from the $k$-th ground device, the achievable throughput is given by
\begin{equation}
\setlength\abovedisplayskip{2pt}
\setlength\belowdisplayskip{2pt}
    C_{t}^{i,k} =
    \begin{cases}
    R_{t}^{i,k}, &\text{if}\, D_{t}^{k}\ge R_{t}^{i,k}\Delta t,\\
    D_{t}^{k}/\Delta t, &\text{otherwise},
    \end{cases}
\end{equation}
which is limited by the information rate $R_{t}^{i,k}$ when the remaining data of the device is sufficient; otherwise, the UAV collects
all remaining data within one time slot duration $\Delta t$.

\subsection{Problem Formulation}
The objective of our algorithm is to design the trajectories for multiple UAVs to maximize the amount of data collected from ground devices within mission time. Utilizing the previously defined UAV mobility model and channel model, this data collection problem can be formulated as the following optimization problem 
\begin{subequations}\label{eq:objective function}
\setlength\abovedisplayskip{5pt}
\setlength\belowdisplayskip{5pt}
\begin{align}
\max _{\times_{i} \mathbf{a}_{t}^{i}} & \sum_{t=0}^{T-1} \sum_{i=1}^{I} \sum_{k=1}^{K} q_{t}^{i,k} C_{t}^{i,k} \Delta t, \tag{\ref{eq:objective function}}  \\
\text { s.t. } & \mathbf{p}_{0}^{i} = \mathbf{p}_{I}, \mathbf{p}_{T}^{i} = \mathbf{p}_{F}, \forall i \in \mathcal{I}, \label{constrain a}\\ 
& h^{i} \neq h^{j}, \forall i \neq j, i,j\in \mathcal{I}, \label{constrain b}\\
& b_{T}^{i} \geq 0, \forall i\in \mathcal{I}, \label{constrain c}\\
& \eqref{eq:UAV_mobility}, \eqref{eq:UAV_battery}, \eqref{eq:device}, \eqref{eq:UAV}, \label{constrain d}
\end{align}
\end{subequations}
where $\times_{i} \mathbf{a}_{t}^{i}$ is the joint action of all the UAVs. \eqref{constrain b} ensures that all the UAVs fly at different altitudes, thereby avoiding collisions between them. \eqref{constrain c} guarantees that all the UAVs can reach the terminal positions with remaining battery at the end of the mission time. This optimization problem is challenging due to its non-convexity and is further complicated because of the unknown channel model and unknown device locations.

%% file: sections/03mdp.tex
\section{Decentralized Partially Observable Markov
Decision Process and QMIX} \label{sec:MDP}
\subsection{Dec-POMDP}
To solve optimization problem \eqref{eq:objective function}, we first reformulate it as a decentralized partially observable Markov decision process (Dec-POMDP) which is defined as a tuple $(\mathcal{I}, \mathcal{S}, \mathcal{A}_{\times}, P, R, \Omega_{\times}, \mathcal{O}, \gamma)$, where $\mathcal{I}$ denotes a set of $I$ agents, $\mathcal{S}$ describes the state space of the environment, $\mathcal{A}_{\times}=\times_{i}\mathcal{A}^{i}$ is the joint action space, according to \eqref{eq:Action space}, $\mathcal{A}^{i} = \mathcal{A}$ is the same for all the UAVs, and $\Omega_{\times}=\times_{i}\Omega^{i}$ is the joint observation space. At each time step, given action $\mathbf{a}^{i}\in\mathcal{A}$ executed by agent $i$, and the joint action $\times_{i}\mathbf{a}^{i}\in \mathcal{A}_{\times}$,  the environment transitions from state $s\in\mathcal{S}$ to next state $s'\in\mathcal{S}$ according to the probability $P(s' \mid s, \times_{i} \mathbf{a}^{i})$. All the agents share the same reward function $R(s,\times_{i}\mathbf{a}^{i},s') = r$.

In a partially observable environment, each agent can only access its local observation $o^{i}\in \Omega^{i}$ according to the observation probability function  $O(\times_{i} o^{i}, s', \times_{i} \mathbf{a}^{i})$, where $\times_{i} o^{i} \in \Omega_{\times}$. Such partial observability is introduced by a limited communication range between UAVs and devices, which is determined by the SNR level of the link. A UAV can communicate with a device if the SNR value of the link between them is greater than or equal to a threshold $\SNR_{thr}$. The same assumption is also applied to the communication between two UAVs. Moreover, we denote the individual local action-observation history of agent $i$ by $\tau_{t}^{i} = (o_{0}^{i}, \mathbf{a}_{0}^{i}, o_{1}^{i},\cdots,\mathbf{a}_{t-1}^{i},o_{t}^{i})$ and the joint action-observation history by $\times_{i} \tau^{i}$. Each agent takes action according to its policy $\pi^{i} \in \Pi^{i}: \Omega^{i} \rightarrow \mathcal{A}^{i}$, and the joint action-value function under joint policy $\pi$ is defined as $Q_{tot}^{\pi}(s, \times_{i} \mathbf{a}^{i}) = \mathbb{E}_{\pi}\left[\sum_{k=0}^{\infty} \gamma^{k} r_{t+k} \mid s_{t}=s, \times_{i} \mathbf{a}_{t}^{i}=\times_{i} \mathbf{a}^{i}\right]$, where $\gamma\in [0, 1]$ is the discount factor. Additionally, the indices of all agents except agent $i$ are denoted by $-i$.

\textbf{Action space.} The action space is defined in \eqref{eq:Action space}. Furthermore, we design a safety controller to only provide feasible actions for the agents to choose from. The safety controller ensures that no collisions with map boundaries occur and all agents reach the destination before their batteries run out by continuously comparing the distance to the destination and remaining battery levels. Note that agents with no remaining battery can only choose the no-op action. Consequently, we denote the feasible action space of agent $i$ at time step $t$ checked by the safety controller by $\mathcal{A}_{t}^{i,sc}$, and the minimum battery required to reach the destination by $b_{t}^{i,sc}$.

\textbf{Observation.} The observation of agent $i$ at time step $t$ is denoted by $o_{t}^{i} = (o_{t,1}^{i}, o_{t,2}^{i}, o_{t,3}^{i}, o_{t,4}^{i})$. To be specific, $o_{t,1}^{i} = \{\mathbf{a}_{t}^{i}\}_{\mathbf{a}_{t}^{i}\in \mathcal{A}_{t}^{i,sc}}$ consists of the feasible actions determined by the safety controller. $o_{t,2}^{i}$ includes the features between agent $i$ and all the devices, i.e., $o_{t,2}^{i} = \{\SNR_{t}^{i,k},\chi_{t}^{i,k},D_{t}^{k},d_{t}^{i, k},d_{t,x}^{i, k},d_{t,y}^{i, k},q_{t}^{i,k}\}_{\forall k\in \mathcal{U}}$, where $\chi_{t}^{i,k}\in \{0,1\}$ is a binary variable, indicating whether device $k$ is reachable by agent $i$. Specifically, if $\SNR_{t}^{i,k} \ge \SNR_{thr}$, then $\chi_{t}^{i,k}=1$; otherwise, $\chi_{t}^{i,k}=0$ and $D_{t}^{k}$ in $o_{t,2}^{i}$ is set to zero. We denote the relative distance between agent $i$ and device $k$ along $x$ and $y$ axis by $d_{t,x}^{i,k}=x_{t}^{i}-x^{k}$ and $d_{t,y}^{i, k}=y_{t}^{i}-y^{k}$, respectively. Similarly, $o_{t,3}^{i}$ includes the features between agent $i$ and other agents, i.e., $o_{t,3}^{i} = \{\SNR_{t}^{i,j},\chi_{t}^{i,j},d_{t}^{i,j},d_{t,x}^{i, j},d_{t,y}^{i,j},b_{t}^{j}\}_{\forall j\in \mathcal{I}^{-i}}$. Notably, if $\chi_{t}^{i,j}=0$, then all elements related to agent $j$ in $o_{t,3}^{i}$ except the SNR value are set to zero.
The last term of the observation includes individual features, i.e., $o_{t,4}^{i} = (b_{t}^{i}, b_{t}^{i, sc})$.

\textbf{State.} The global state contains the information of all the UAVs and devices, regardless of the SNR value, i.e., $s_{t} = (s_{t,1}, s_{t,2})$, where $s_{t,1} = \{b_{t}^{i}, b_{t}^{i, sc}, x_{t}^{i}, y_{t}^{i}, \text{done}\}_{i\in \mathcal{I}}$ includes the features of all the UAVs, $\text{done}\in \{0,1\}$ indicates whether the UAV's battery is fully depleted, and $s_{t,2} = \{D_{t}^{k}, x^{k}, y^{k}\}_{k\in \mathcal{U}}$ contains the features of all the devices. This global state is only available during centralized training.

\textbf{Reward.} The joint reward received at each time step $t$ is defined as the amount of data collected by all the UAVs which is given by
\begin{equation}
\setlength\abovedisplayskip{2pt}
\setlength\belowdisplayskip{2pt}
    r_{t} = \sum_{i=1}^{I} \sum_{k=1}^{K} q_{t}^{i,k}  C_{t}^{i,k} \Delta t.
    \label{eq:reward}
\end{equation}
\vspace{-2.2em}
\subsection{QMIX}
To solve the Dec-POMDP, we adopt the popular QMIX \cite{rashid2020monotonic} algorithm, which non-linearly factorizes the joint action-value function using a mixing network with leverage of the global state information, as described by
{
\medmuskip=0.25mu
\thinmuskip=0.25mu
\thickmuskip=0.25mu
\begin{equation}
\setlength\abovedisplayskip{5pt}
\setlength\belowdisplayskip{5pt}
    Q_{t o t}(\times_{i}\tau^{i}, \times_{i}\mathbf{a}^{i}) = \text{Mix}(s, Q_{1}(\tau^{1}, \mathbf{a}^{1}),\cdots,Q_{I}(\tau^{I}, \mathbf{a}^{I});\theta),
\end{equation}
}where $Q_{i}(\tau^{i}, \mathbf{a}^{i})$ is the individual action-value function conditioned on the local observation-action history $\tau^{i}$ and action $\mathbf{a}^{i}$, and $\theta$ denotes the parameters of the QMIX model.
We randomly sample a batch of $B$ episodes from the replay buffer to train the QMIX model by minimizing the following loss
\begin{equation}
\mathcal{L}(\theta)=\sum_{b=1}^{B}\sum_{t=0}^{T-1} \left(y_{b,t}^{tot}-Q_{tot}(\times_{i}\tau_{t}^{i}, \times_{i}\mathbf{a}_{t}^{i}, s_{t} ; \theta)\right)^{2}, \label{eq:QMIX loss}
\end{equation}
where $y_{b,t}^{t o t}=r_{b,t}+\gamma \max _{\times_{i}{\mathbf{a}^{i}}} Q_{t o t}(\times_{i}{\tau_{t+1}^{i}}, \times_{i}{\mathbf{a}^{i}}, s_{t+1}; \Bar{\theta})$ denotes the temporal-difference (TD) target at time step $t$ of the $b$-th episode, $\Bar{\theta}$ denotes the target network parameters.

%% file: sections/04model.tex
\section{Model-aided FedQMIX}
Employing MARL algorithms directly in real-world scenarios is usually impractical since collecting large amounts of training data on physical systems is expensive and can lead to potential safety concerns \cite{dulac2019challenges}. To tackle these challenges, we propose a model-aided federated MARL algorithm named \emph{model-aided FedQMIX} for planning multi-UAV trajectories in data harvesting missions while significantly reducing the requirement for real-world training data samples.

Our assumption about the environment setting lies between agnostic and fully informed. On the one hand, we do not assume prior knowledge of the wireless channel characteristics, which must be learned by the UAVs, nor do we presume knowledge of all devices' positions, which must be estimated to construct the simulated environment. On the other hand, a 3D map of the environment and position information of only a subset of devices (anchor devices) are known, which gives the UAVs the necessary knowledge to localize the devices with unknown positions and develop an accurate simulated model.

The proposed algorithm alternates between two parts: 1) learning a simulated environment from real-world measurements, and 2) training the QMIX model for trajectory planning via federated learning in the simulated environment. Specifically, UAVs collect measurements from ground devices in the real world, which are then used to learn the radio channel model and determine the unknown device locations, as introduced in Section IV-A. The learned information, combined with a 3D map, is used to build a simulated environment for each UAV agent.
Subsequently, we train a global QMIX model by exploiting all the UAVs' resources via federated learning in the simulated environment, as detailed in Section \ref{sec:FedQMIX}.

\subsection{Environment Learning}\label{sec:Environment learning}
Akin to \cite{esrafilian2021model} but considering a multi-UAV setting, we employ a neural network to learn the radio channel and use particle swarm optimization (PSO) along with the learned radio channel to estimate unknown device locations.

During each real-world deployment and in addition to data collection, each UAV measures the channel gain between itself and the devices. Upon episode completion, all UAVs transmit the gathered measurements to a central server or a specified UAV for environment model learning. The channel gain between the $i$-th UAV and the $k$-th device at time step $t$ can be defined as a function $\psi$ with parameters $\vartheta$, as follows
\begin{equation}\label{eq:RSS function}
\setlength\abovedisplayskip{5pt}
\setlength\belowdisplayskip{5pt}
g_{t}^{i,k}  {=}
\begin{cases} 
    \psi_{\vartheta}(d_{t}^{i, k}, \phi_{t}^{i, k}, \omega_{t}^{i, k}=1)+\eta_{\text{LoS}},       & \small{\text{if} \text{ LoS}},\\
    \psi_{\vartheta}(d_{t}^{i, k}, \phi_{t}^{i, k}, \omega_{t}^{i, k}=0)+\eta_{\text{NLoS}},       & \small{\text{if} \text{ NLoS}},
  \end{cases}
\end{equation} 
where $\phi_{t}^{i,k}=\arcsin ({h^{i}}/{{d}_{t}^{i,k}})$ is the elevation angle. The binary variable $\omega_{t}^{i, k}\in\{\text{0, 1}\} $ indicates whether a measurement falls into LoS category for $\omega_{t}^{i, k} = 1$ or NLoS otherwise. The shadowing effect $\eta_{z}$ is characterized as $\mathcal{N}(0,\sigma_{z}^2)$. Note that both the function $\psi$ and the parameters $\vartheta$ are unknown and must be learned. Each measurement in \eqref{eq:RSS function} can be modeled as
$p(g_{t}^{i,k})=(f_{t,\text{LoS}}^{i,k})^{\omega_{t}^{i, k}} (f_{t,\text{NLoS}}^{i,k})^{(1-\omega_{t}^{i,k})}$, where $f_{t,z}^{i,k} = \mathcal{N}(\psi_{\vartheta}(d_{t}^{i, k}, \phi_{t}^{i, k}, \omega_{t}^{i, k}), \sigma_{z}^{2})$. 
Akin to \cite{esrafilian2021model}, the negative log-likelihood of the measurements is given by
\begin{equation}
\begin{aligned}
\mathcal{L} &= \log \left(\frac{\sigma_{\text{LoS}}^{2}}{\sigma_{\text{NLoS}}^{2}}\right) \sum_{t=0}^{T} \sum_{i=1}^{I} \sum_{k=1}^{K} \omega_{t}^{i,k} \\
& + \sum_{t=0}^{T} \sum_{i=1}^{I} \sum_{k=1}^{K}  \frac{\omega_{t}^{i,k}}{\sigma_{\text{LoS}}^{2}}\left|g_{t}^{i,k}-\psi_{{\vartheta}}(d_{t}^{i,k}, \phi_{t}^{i,k}, \omega_{t}^{i,k})\right|^{2} \\
& + \sum_{t=0}^{T} \sum_{i=1}^{I} \sum_{k=1}^{K} \frac{1-\omega_{t}^{i,k}}{\sigma_{\text{NLoS}}^{2}}\left|g_{t}^{i,k}-\psi_{{\vartheta}}(d_{t}^{i,k}, \phi_{t}^{i,k}, \omega_{t}^{i,k})\right|^{2} .
\end{aligned}
\end{equation}
Consequently, the problem of learning the channel model and estimating the device locations can be transformed into solving the optimization problem
\begin{equation}
\begin{split}
\min_{\substack{\omega_{t}^{i,k}, \mathbf{u}^{k}, \forall t, \forall i, \forall k \\
\psi(\cdot), \vartheta}} & \mathcal{L}, \\
\text { s.t. } & \omega_{t}^{i,k} \in\{0,1\}, \forall t, \forall i, \forall k \text {. }
\end{split}\label{eq:log-likelihood}
\end{equation}
The above optimization problem is non-convex and hard to solve directly. Therefore, we decompose \eqref{eq:log-likelihood} into two sub-problems: i) radio channel learning, and ii) device localization. In phase one, the radio channel is learned by utilizing the known locations of anchor devices, and in the second phase, the learned channel from the previous step is combined with a PSO algorithm and the 3D map to localize the unknown devices. Due to the limited space, we refer to \cite{esrafilian2021model} for details.

Having estimated the unknown device locations and learned the radio channel, we utilize this information along with the map to build a simulated environment for training the MARL algorithm to design the trajectories for the UAVs.
    
\subsection{QMIX Model Training via Federated Learning} \label{sec:FedQMIX}

Adapting the idea of federated learning to make efficient use of all UAVs' computational resources and accelerate the learning process, we create a digital replica of the environment locally at each UAV, also including virtual instances of all other UAVs. By running the same MARL algorithm simultaneously on each UAV and consolidating the trained models using federated learning, we obtain a global QMIX model.

While training locally in the simulated environments at each UAV, virtual agents may take different actions even if they have the same observations because of the random component in the $\epsilon$-greedy action selection strategy, leading to diverse sets of state transitions in individual replay buffers. Federated learning can help to capitalize on the diversity of experiences in all UAVs' replay buffers by periodically consolidating locally learned QMIX models. This allows us to make efficient use of all available training experiences while avoiding the excessive communication overhead of centralized learning.

Specifically, each UAV $i$ trains its QMIX network parameters $\theta^{i}$ in its own simulated environment. Every $N_{freq}$ episodes, all the UAVs periodically send their trained model to an aggregator, which computes a global model by averaging over local models as  $\theta = \frac{1}{I} \sum_{i\in\mathcal{I}} \theta^i$.
The aggregator then sends the global model back to each UAV, where training in their respective local simulated environments continues.

\subsection{Algorithm}
The model-aided FedQMIX algorithm is summarized in Algorithm \ref{alg:Ma-FLQMIX} and consists of three steps: 1) The UAVs employ the learned policy to generate trajectories for collecting data and measurements in the real world; 2) The acquired measurements are used to learn the radio channel and estimate unknown device locations, which are utilized to build the simulated environment; 3) The UAV agents train the QMIX model using federated learning in their respective simulations for a predefined number of episodes. The algorithm terminates after carrying out $E_{max}$ real-world experiments.

\SetAlgoSkip{}
\setlength{\algomargin}{0.1em}

\begin{algorithm}[]
{\small
    \caption{Model-aided FedQMIX}\label{alg:Ma-FLQMIX}
    \begin{algorithmic}[1]
    \STATE {Initialize a set of $I$ UAVs, replay buffers $\mathcal{B}^{i}$, the parameters $\theta$ of QMIX at the aggregator, local QMIX parameters $\theta^{i}=\theta$, target network parameters $\Bar{\theta}^{i}=\theta^{i}$, target network update period $N_{\textit{target}}$, aggregation period $N_{\textit{freq}}$.}
    \FOR{$e = 0,1, \cdots, E_{max}-1$}
        \STATE {\textbf{1) Real-world experiment:}}
        \STATE{UAVs use the policy derived from step 3) to plan trajectories for data collection while also gathering measurements.}
      \STATE{\textbf{2) Learn the environment as described in Section \ref{sec:Environment learning} }}
        \STATE{\textbf{3) Simulated-world experiment:}}
        \FOR{$episode = 0, 1, \cdots, N-1$}
        \FOR{Each UAV $i \in \mathcal{I}$ in parallel}
            \STATE{$t=0$, initialize state ${s}_{0}$}  
            \WHILE{$b_{t}^{j} \ge 0, \forall j = 1,2, \dots,I$}
                \FOR{each simulated agent $j = 1,2,\dots,I$}
                    \STATE{${\tau}_{t}^{j} = {\tau}_{t-1}^{j} \cup \lbrace{({o}_{t}^{j}, {\mathbf{a}}_{t-1}^{j})}\rbrace$}
                    \STATE{Choose action with $\epsilon$-greedy policy, i.e., 
                    $\mathbf{{a}}_{t}^{j}=\begin{cases}
                    \text{randomly select from } \mathcal{A}_{t}^{j,sc}, & \text{w.p. } \epsilon \\
                    \arg \max_{\mathbf{a}_{t}^{j} \in \mathcal{A}_{t}^{j,sc}} Q_{j}({\tau}_{t}^{j}, \mathbf{a}_{t}^{j}), & \text{w.p. } 1-\epsilon
                    \end{cases}$}
                \ENDFOR
                \STATE{Take joint action $\times_{j}{\mathbf{a}}_{t}^{j}$, observe $\times_{j}{o}_{t+1}^{j}$, get reward ${r}_{t}$ and next state ${s}_{t+1}$}
                \STATE{Store $({s}_{t},\times_{j}{o}_{t}^{j}, \times_{j}{\mathbf{a}}_{t}^{j},{r}_{t},{s}_{t+1}, \times_{j}{o}_{t+1}^{j})$ in $\mathcal{B}^{i}$}
                \STATE{$t = t + 1$}
            \ENDWHILE
            \STATE{Randomly sample a batch of $B$ episodes from $\mathcal{B}^{i}$}
            \FOR{each time step $t$ in each episode in the batch}
                \STATE{$Q_{tot}=\text{Mix}({s}_{t},Q_{1}({\tau}_{t}^{1},{\mathbf{a}}_{t}^{1}), \cdots, Q_{I}({\tau}_{t}^{I}, {\mathbf{a}}_{t}^{I});\theta^{i})$}
                \STATE{Calculate target $Q_{tot}$ using target network $\Bar{\theta}^{i}$}
            \ENDFOR
            \STATE{$\theta^i \leftarrow \theta^i - \alpha \nabla \mathcal{L}(\theta^i) $ w.r.t. $\theta^i$ using Eq. \eqref{eq:QMIX loss}}
            \IF{$mod(episode, N_{\textit{target}}) = 0$}
                \STATE{Reset $\Bar{\theta}^{i} = \theta^{i}$}
            \ENDIF
        \ENDFOR
        \IF{$mod(episode, N_{\textit{freq}}) = 0$}
            \STATE{Update $\theta = \frac{1}{I} \sum_{i\in\mathcal{I}} \theta^i$ and set $\theta^i \leftarrow \theta, \forall i\in\mathcal{I}$}
        \ENDIF
    \ENDFOR
    \ENDFOR
    \end{algorithmic}
}
\end{algorithm}

%% file: sections/05results.tex
\addtolength{\topmargin}{0.05in}
\section{Numerical Results}\label{sec:simulations}
\begin{figure}[t]
\setlength\abovecaptionskip{.1cm}
\setlength\belowcaptionskip{-.5cm}
\centering
\subfigure[Return-Base Map (RBM)]{
\begin{minipage}[t]{0.8\linewidth}
\centering
\includegraphics[width=\linewidth]{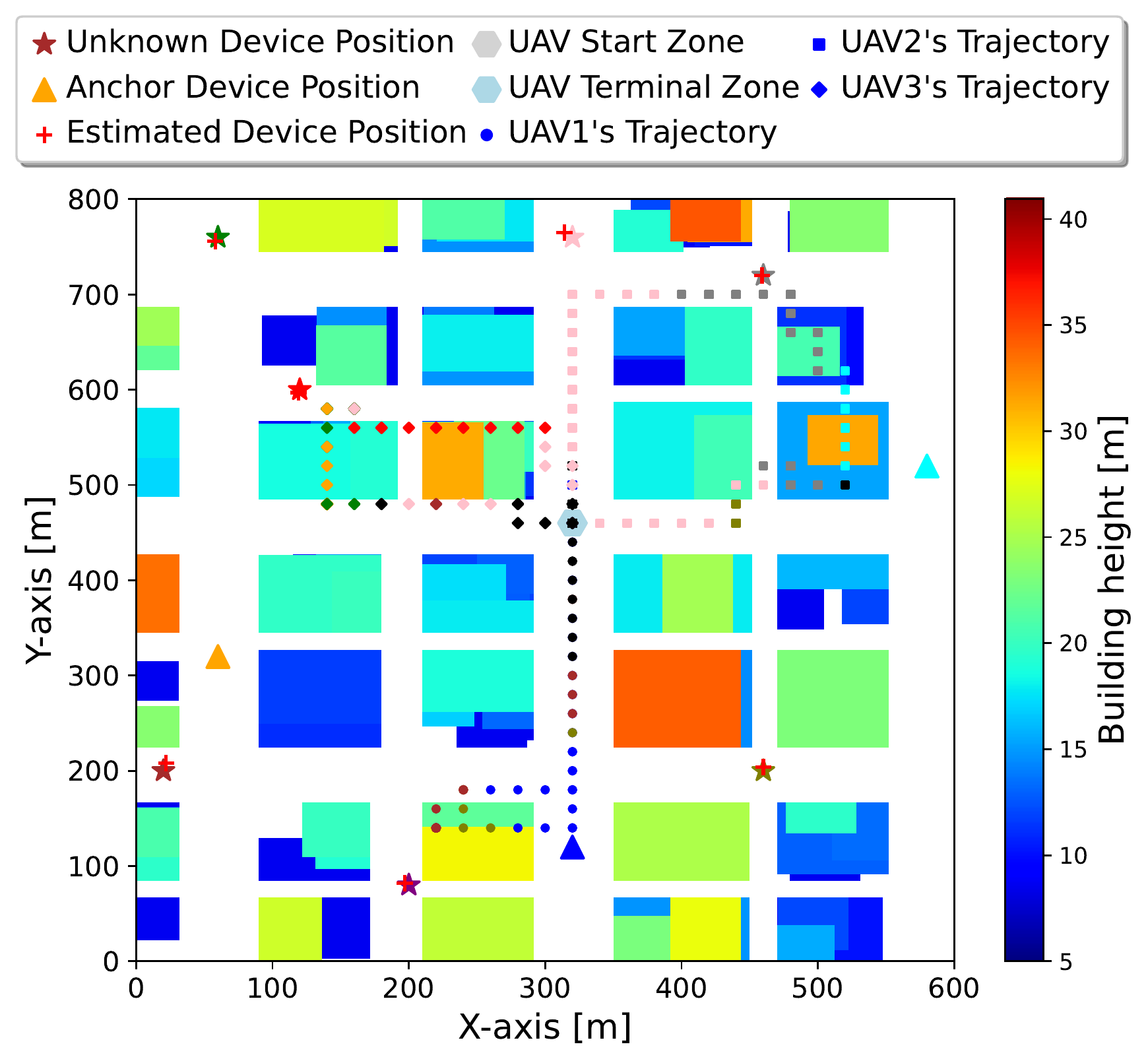}
\vspace{-0.75em}
\label{fig:RBM_trj}
\end{minipage}%
}
\subfigure[Reach-Destination Map (RDM)]{
\begin{minipage}[t]{0.8\linewidth}
\centering
\includegraphics[width=\linewidth]{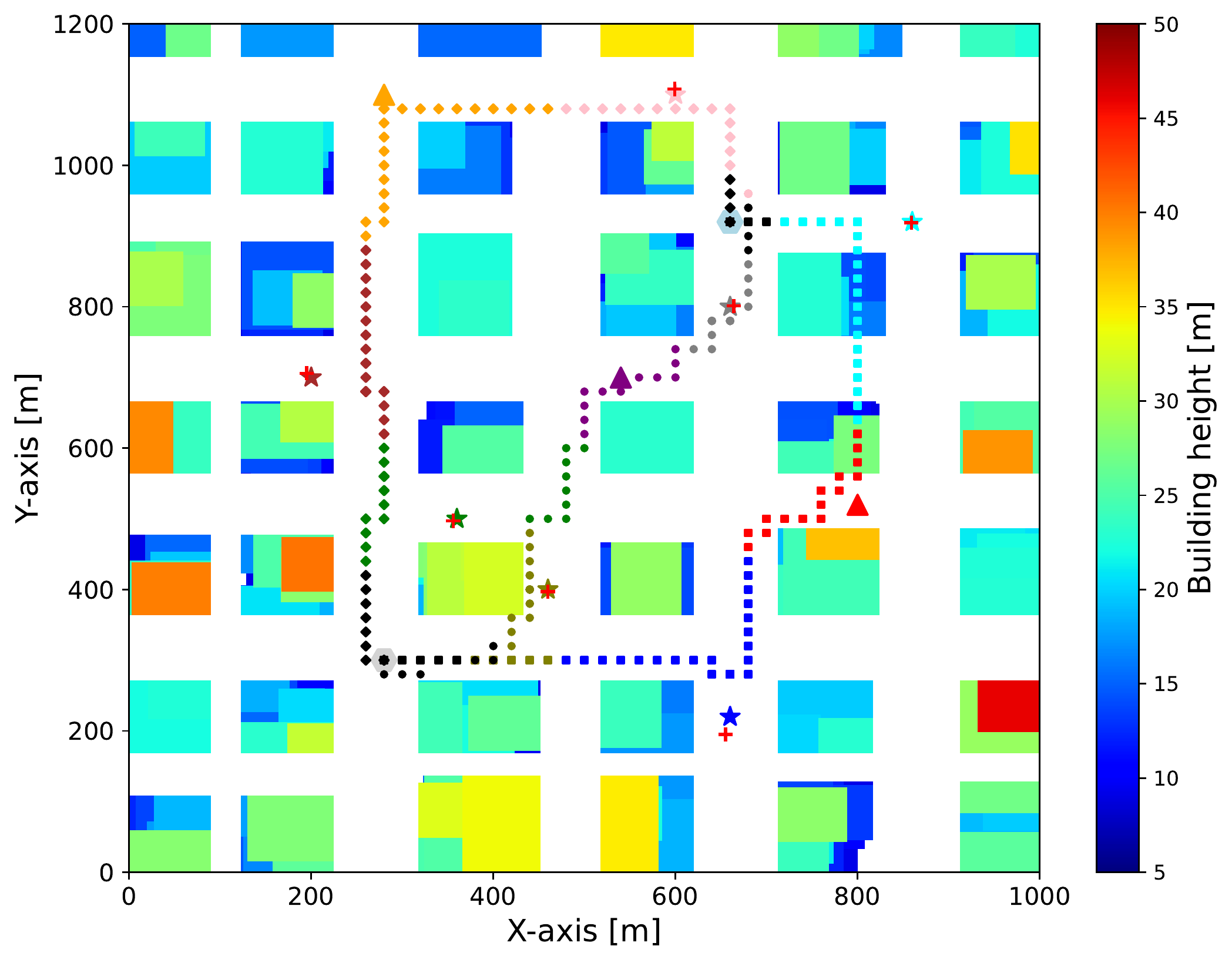}
\vspace{-0.75em}
\label{fig:RDM_trj}
\end{minipage}%
}%
\centering
\caption{Example trajectories, where the UAV's trajectory color corresponds to the device it is collecting data from, while black indicates no data collection. Anchor IoT devices are represented as triangles and devices with unknown locations as stars. a) RBM: $K\!=\!10$ devices with $D_{init}^{k}\!=\!16000$ initial data units to be picked up by three UAVs with $b_{0}^{i}\!=\!60$ initial battery units. b) RDM: $K\!=\!10$ devices with $D_{init}^{k}\!=\!20000$ initial data units to be picked up by three UAVs with $b_{0}^{i}\!=\!80$ initial battery units.}
\label{fig:trjs}
\end{figure}

We consider a 3D urban environment composed of city blocks with buildings of different heights. We design two types of map: the Return-Base Map (RBM) and the Reach-Destination Map (RDM), the top views of which are illustrated in Fig. \ref{fig:RBM_trj} and Fig. \ref{fig:RDM_trj}, respectively. The RBM covers an area of $600\si{\metre} \times 800\si{\metre}$ with the same start and terminal positions for the UAVs located at the center of the map, while the RDM extends over a larger area of $1000\si{\metre} \times 1200\si{\metre}$, where the UAVs are required to navigate from the start position situated at the lower left corner to the destination located at the upper right corner. The flying altitudes of the UAVs are set as 55m, 60m, and 65m, respectively. We adopt the same QMIX hyper-parameters as in \cite{rashid2020monotonic}, except that we use the Adam optimizer with a learning rate of $5 \times 10^{-4}$. True propagation parameters are chosen similar to \cite{esrafilian2021model}. The federated learning aggregation period is set as $N_{freq} = 50$ and a total of $E_{max} = 30$ real-world episodes are executed during training.
After completing a real-world episode, each UAV undergoes local training for $N=1000$ episodes in its simulated environment. This helps to reduce the need for expensive real-world experiments.

Fig. \ref{fig:trjs} shows two example trajectories generated by the model-aided FedQMIX algorithm for three UAVs. 
On both maps, the UAVs learn to efficiently divide the entire map into sub-regions that are then assigned to individual UAVs, without the need for centralized coordination. Total collected data is thereby maximized and energy wastage by UAVs congregating is avoided, which demonstrates the effectiveness of the learned cooperative behavior. In Fig. \ref{fig:trjs}, the red crosses indicate the estimated locations of unknown devices in the final episode, which are closely aligned with the actual device positions and showcase the accuracy of the learned environment model.

\begin{figure}[t]
\setlength\abovecaptionskip{0.1cm}
\setlength\belowcaptionskip{-0.5cm}
\centering
\subfigure[Return-Base Map (RBM)]{
\begin{minipage}[t]{0.48\linewidth}
\centering
\includegraphics[width=\linewidth]{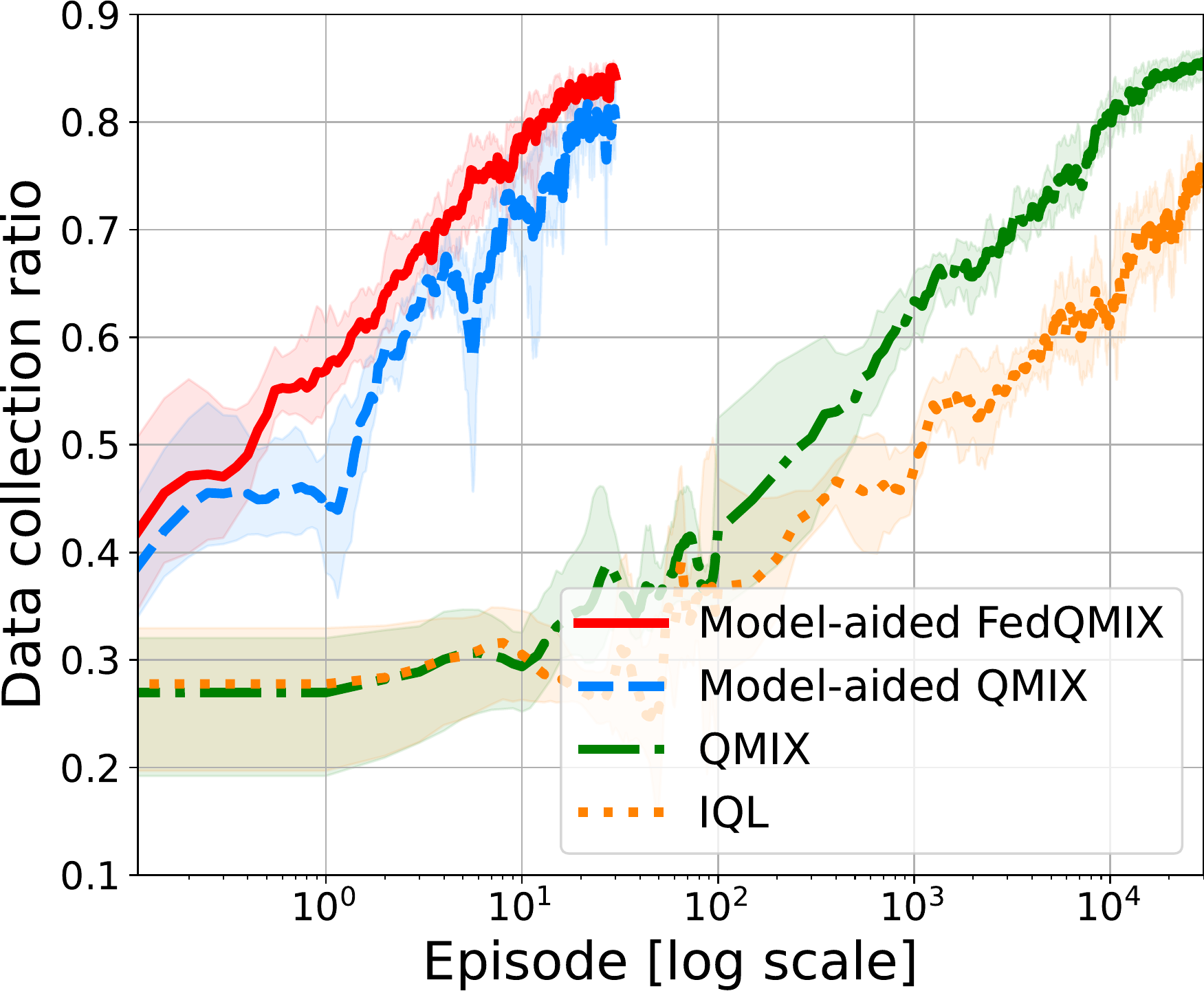}
\label{fig:RBM_pfm}
\end{minipage}%
}%
\subfigure[Reach-Destination Map (RDM)]{
\begin{minipage}[t]{0.48\linewidth}
\centering
\includegraphics[width=\linewidth]{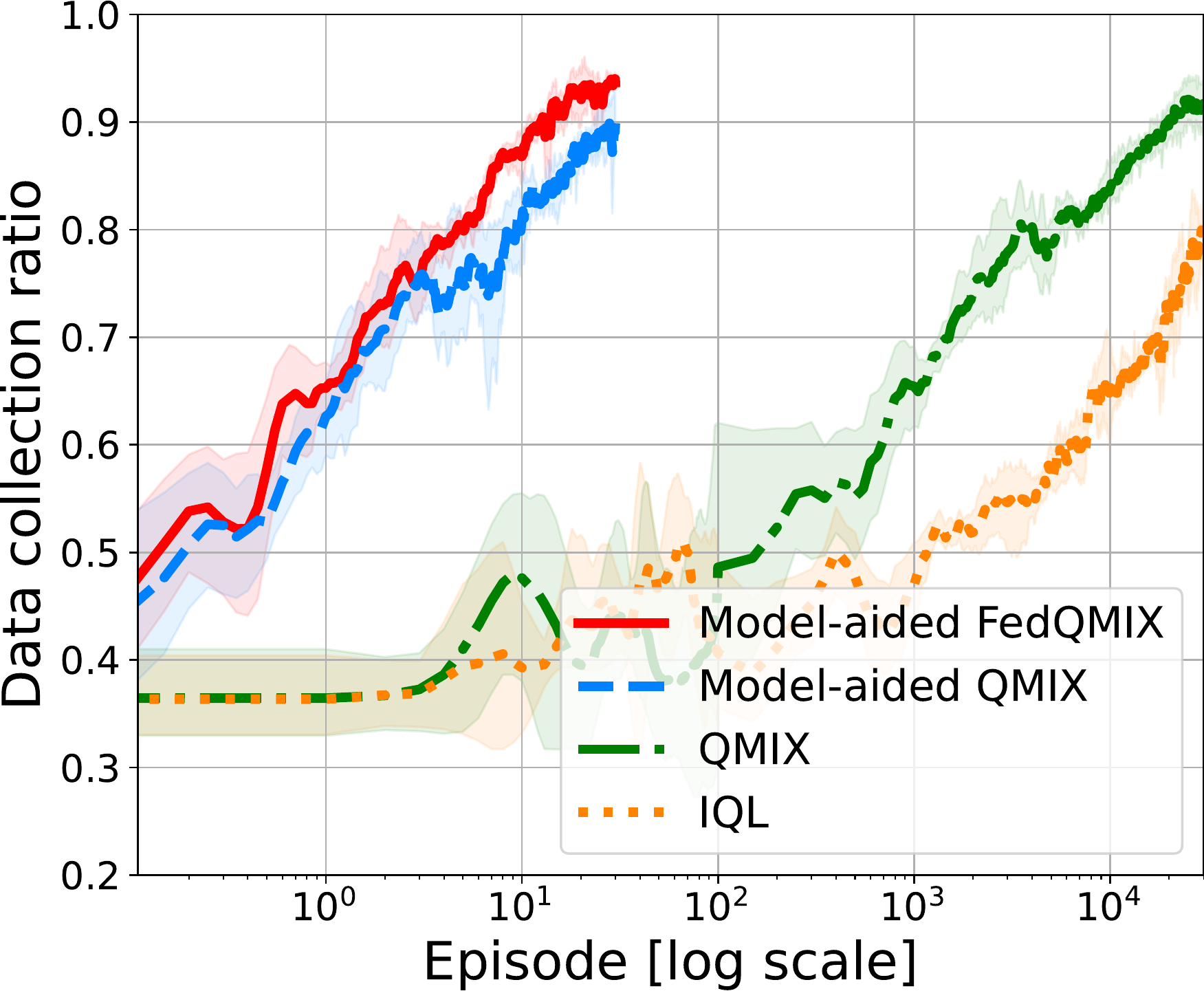}
\label{fig:RDM_pfm}
\end{minipage}%
}%
\centering
\caption{Performance comparison of proposed model-aided FedQMIX and baseline algorithms. The x-axis indicates real-world training episodes on a logarithmic scale. Results are averaged over three random runs.}
\label{fig:Performance comparisons}
\end{figure}

We compare our proposed algorithm with several benchmarks including the conventional QMIX algorithm without model learning, and fully decentralized independent Q-learning (IQL). 
Besides, we also train the QMIX model without federated learning, referred to as model-aided QMIX. In this approach, the training only occurs in one of the UAVs' simulated environments. Once the training is finished, the learned policy is shared with other UAVs.
The performance comparison results of both maps are shown in Fig. \ref{fig:Performance comparisons}. The IQL approach considerably underperforms QMIX due to the non-stationarity issue in independent learning methods \cite{zhang2021multi}. The model-aided QMIX algorithm demonstrates a notable reduction in the need for real-world experiences while attaining comparable performance levels to the baseline QMIX algorithm. Significantly, our proposed model-aided FedQMIX algorithm shows superior performance compared with other approaches. It achieves the same performance as QMIX trained with real-world samples while requiring around three magnitudes less real-world data. Additionally, in comparison with model-aided QMIX, it converges faster and achieves better performance within the equivalent training timeframe. 
This can be attributed to the adoption of federated learning in training the QMIX model, which exploits the information from all UAVs' trained models, resulting in faster convergence and distributed computation load across UAVs.

